\documentclass[letterpaper, 10 pt, conference]{ieeeconf}  

\IEEEoverridecommandlockouts                              

\usepackage{times} 
\usepackage{amsmath} 
\usepackage{amssymb}  
\usepackage{hyperref}
\usepackage{booktabs}
\usepackage{url}
\usepackage{mathtools}
\usepackage[ruled,vlined]{algorithm2e}

\title{\LARGE \bf
Improving Generalization of Transfer Learning Across Domains Using Spatio-Temporal Features in Autonomous Driving
}

\author{Shivam Akhauri$^{2}$, Laura Y. Zheng$^{1}$, Tom Goldstein$^{1}$
        and Ming C. Lin$^{1,2}$ \\
        \thanks{The authors are with 
$^{1}$Department of Computer Science and
$^{2}$Maryland Robotics Center,
         University of Maryland at College Park, MD, U.S.A.
        E-mail: \{sakhauri,lyzheng,tomg,lin\}@umd.edu} %
       \url{https://gamma.umd.edu/stltransfer}
       }
       
\usepackage{etoolbox}
\makeatletter
\patchcmd{\@makecaption}
  {\scshape}
  {}
  {}
  {}
\makeatother

\begin{document}

\maketitle
\thispagestyle{empty}
\pagestyle{empty}

\begin{abstract}
Practical learning-based autonomous driving models must be capable of generalizing learned behaviors from simulated to real domains, and from training data to unseen domains with unusual image properties. In this paper, we investigate transfer learning methods that achieve robustness to domain shifts by taking advantage of the invariance of spatio-temporal features across domains. 
In this paper, we propose a transfer learning method to improve generalization across domains via transfer of spatio-temporal features and salient data augmentation. Our model uses a CNN-LSTM network with Inception modules for image feature extraction. Our method runs in two phases: Phase 1 involves training on source domain data, while Phase 2 performs training on target domain data that has been supplemented by feature maps generated using the Phase 1 model.
%
%
Our model significantly improves performance in unseen test cases for both simulation-to-simulation transfer as well as simulation-to-real transfer by up to +37.3\% in test accuracy and up to +40.8\% in steering 
angle prediction, compared to other SOTA methods across multiple datasets.

\end{abstract}


\section{INTRODUCTION}

There is an abundance of freely available datasets for training autonomous driving systems, both from real-world and simulated environments.  While the environments and image properties of such datasets may differ from a user's target domain, transfer learning enables systems to benefit from a range of existing data sources, including simulated data that captures events and scenarios that are difficult to observe in the real world.  
Virtually all driving datasets share high-level visual elements, including lane markings, a horizon and sky, and complex scenes with cars, signs, lights, and trees. An effective machine learning system should benefit from the fact that the co-occurrence of these objects is nearly the same across datasets, even if the pixel representations are different.

End-to-end driving frameworks based on behavior cloning, such as the pioneering work by Bojarski et al.~\cite{nvidia} among others~\cite{chen2017end, 9165167, 7410669, 8733812}, primarily use Convolutional Neural Networks (CNN) to perform tasks related to lane keeping or full-vehicle control based on offline data provided by an expert driver. Such end-to-end systems should be benefit from transfer learning, provided one can mitigate the effects of distribution shift and dataset bias described by Codevilla et al.~\cite{codevilla2018end}.

In addition to spatial information provided by the CNN, a driving model can also benefit greatly from temporal information. The notion of spatio-temporal end-to-end driving has already been explored, especially with videos~\cite{Xu_2017_CVPR}. However, there has been little discussion on utilizing {\em learned spatio-temporal features} in transfer learning. Transfer learning with spatio-temporal features is challenging due to the temporal information often being too disparate in source and target domains. In the case for autonomous driving, this obstacle can be mitigated due to the natural movement of cars in traffic. While there may be differences across datasets in scenery, animals, right or left handedness, traffic signs, etc., these differences are on a smaller granularity compared to the ubiquitous consistency of lane markings, and the universal physics that governs traffic flow. Thus, temporal dynamics is a natural target for improvement in transfer learning due to the consistency of how objects move across frames. Spatio-temporal information captured from driving data in any domain will be more domain-agnostic than solely spatial information alone. 

We present a transfer learning method for autonomous driving that utilizes spatio-temporal features during training to improve cross-domain generalization. Our end-to-end model involves a CNN+LSTM architecture, from which we extract the learned weights in the CNN layers~\cite{nvidia} and the hidden state embeddings that capture the spatial-temporal features from the LSTM layers. 

Transfer learning is conducted in two phases. Phase 1 involves straightforward training of the CNN+LSTM model on a source domain. We use simulation image data from Town 3 in the CARLA simulator~\cite{CARLA}. 
Before starting Phase 2 training, we generate saliency, gradient, and edge maps using classifiers trained on Phase 1 data. Saliency and gradient maps are obtained using an external network trained on a larger number of classes. This incorporates information on the data-level, while cross-domain spatio-temporal information is transferred on the model level. 
During Phase 2, CNN weights and LSTM hidden states from Phase 1, representing spatial and temporal features respectively, are transferred to a new model. This model is trained on target domain data (e.g. CARLA Town 1) with the salient data augmentation (including saliency, gradient, and edge maps). 

Our model shows significant improvements in classification test accuracy of up to {\bf 37.29\%} and up to  {\bf +40.8\%}  in  steering angle  prediction  for unseen environments compared to CNN-only models~\cite{akhaurizhenglin2020}. Unseen test domains include other simulation data with visual differences and real-world data, which are never introduced to the model during phase one training. We also perform an ablation study on the effect of spatio-temporal features in learning, both on the architectural level and on the data augmentation level.

In summary, the main contributions of this work on {\em using spatial-temporal features at both the network and the data levels to improve transfer learning} include: 
\begin{enumerate}
    \item A 2-Phase transfer learning method involving transfer of LSTM hidden states embeddings and CNN weights~\cite{akhaurizhenglin2020} via weight initialization.
    \item Salient data generation using salient features extracted from external models pre-trained on diverse datasets, to hint towards the context of the source and the target domain data.
\end{enumerate}

\section{Related Work}

\subsection{Spatio-temporal Features in Driving}
Recent works have leveraged spatio-temporal features for both perception and control tasks in autonomous driving. Recurrent Neural Networks (RNNs) and Long-Short Term Memory (LSTM) architectures are the most popular method in this application of spatio-temporal information ~\cite{lstm-hochreiter}. In terms of RNNs, Haavaldsen et al. achieve improvements on the steering control task using CNN-RNN model~\cite{haavaldsen2019autonomous}. This model was qualitatively observed to identify object dynamics better in scenes with sudden movement changes. Other works have used CNN+LSTMs in end-to-end control models~\cite{chen2019cnnlstmdriving,bai2018deep}. We use a CNN+LSTM in our method to further extend on this architecture in the context of transfer learning. 

Most notably in the natural language processing domain, transformer models have become a popular alternative to LSTMs in terms of leveraging spatio-temporal features. Transformers in context of autonomous driving are a relatively new method~\cite{vaswani2017attention_transformers}. ~\cite{eraqi2017end} incorporate the best of both worlds by applying encoder-decoder models on top of a base CNN+LSTM model for steering angle prediction. ~\cite{yuan2021temporal} proposed a new transformer, the Temporal-Channel Transformer (TCTR), for object detection from 3D lidar input. 
While transformers seem like a promising new alternative to LSTMs in terms of performance and interpretability, there are still many implicit obstacles in applying them to autonomous driving control tasks. The benefit of transformers become outweighed by the $O(n^2)$ computation complexity of each layer, where $n$ is the data sequence length. In real-time control tasks, this is an expensive obstacle. Thus, we focus transfer learning efforts on CNN+LSTM architectures in order to maintain relevancy to other methods. 

\subsection{Transfer Learning for Driving}
    Transfer learning can be useful in autonomous driving in transferring information learned in one environment to another, or even in one domain to another. From a survey of transfer learning by Pan et al.~\cite{survey-transfer-learning}, we can classify autonomous driving transfer learning as mostly transductive transfer learning~\cite{Arnold_Nallapati_Cohen}. Transductive transfer learning pertains to the scenario when labeled source domain data is readily available, but target domain data is not. Weiss et al. formally define transfer learning as well as current state of the art in a more recent survey from 2016~\cite{TransferLearningSurvey}. 
    
    There are various tasks researched in the scope of transfer learning and autonomous driving. ~\cite{isele2017transferring} explore transferring knowledge learned from simulated intersections to real-world intersections. \cite{VirtualRealWilderness} applies transfer learning in wilderness traversal, where a lack of real-world wilderness prevents learning from real world data directly. Since the same problem affects autonomous driving in cases such as accident data, transfer learning can be beneficial in transferring learned variables to the real world. Virtual-to-real transfer learning can also be used for reinforcement learning in autonomous driving, as shown by~\cite{VirtualToRealRL}. Likewise, transfer learning has also been shown to improve performance in steering tasks using imitation learning, as shown by~\cite{akhaurizhenglin2020}, which is a transfer learning enhanced framework based on the end-to-end learning model by Bojarski et al.~\cite{nvidia}. In our framework, we use a transfer learning framework to transfer learned information on critical decision making components across environments, which allows for better generalization across domains. Our goal is to learn more relevant and generalizable features from the source domain and minimize the fine tuning needed to adapt to target domains; we achieve this through transfer of spatio-temporal features in transfer learning.

    \begin{figure*}[th!]
    \centering
    \includegraphics[width=12cm,height=6cm]{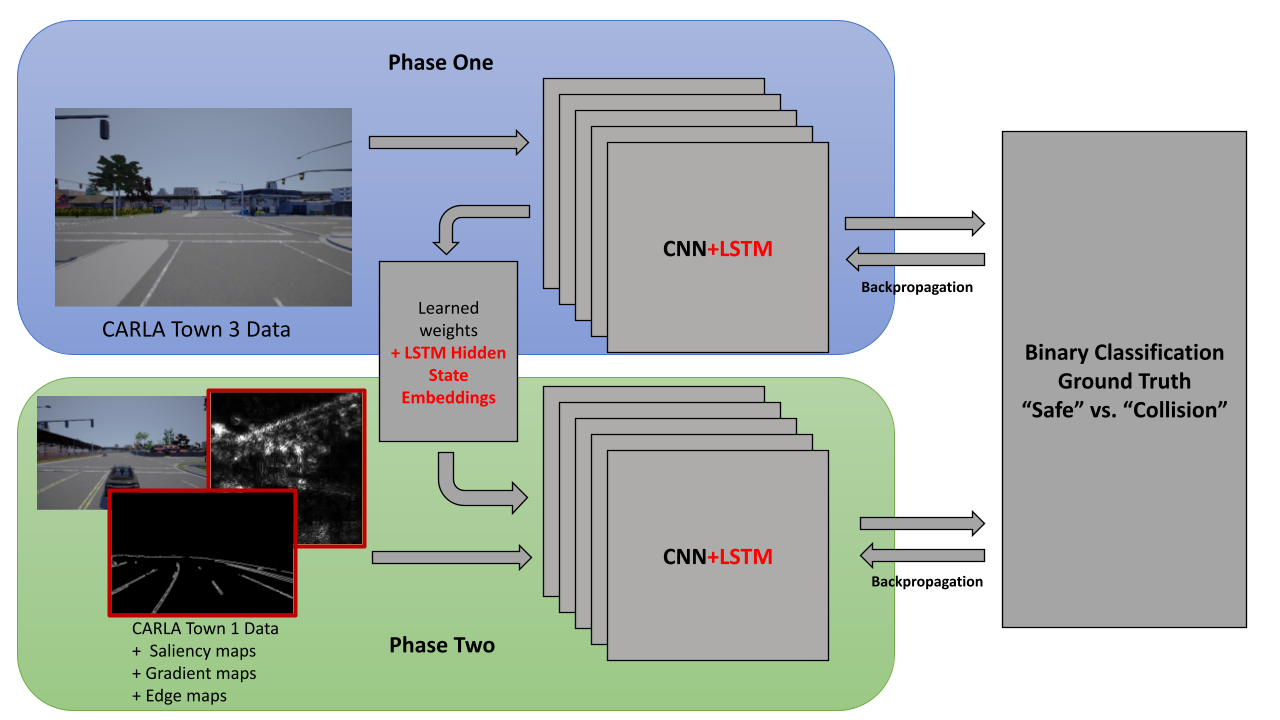}
    \vspace*{-1em}
    \caption{{\bf Transfer learning training framework}. 
    Phase one, highlighted in blue, represents standard training of the CNN+LSTM network with CARLA Town 3 data. Learned weights from CNN and LSTM hidden state embeddings are transferred to Phase Two, highlighted in green. Phase Two training data from CARLA Town 1 is also complemented with Town 1 saliency, gradient, and edge maps generated from preliminary training of Phase Two. 
    Contributions to earlier CNN-based transfer learning frameworks are highlighted in red. 
    }
    \vspace*{-2em}
    \label{fig:experiment-diagram}
\end{figure*}

\section{Enhanced Transfer Learning}
\label{sec:TXF}


Transfer learning can be useful for vision-based autonomous driving due to the abundance and richness of public datasets. However, because of the numerous datasets collected under different scenarios and circumstances, driving models are subject to dataset bias, as mentioned by ~\cite{codevilla2018end}. Transfer of useful driving information may be limited by spatial information. Models may be biased towards visual differences, such as quality, coloring, brightness, and textures, neglecting spatial-temporal dynamics that we will address through transfer learning in this section.

\subsection{Hypothesis}

Conventional transfer learning for vision-based driving generally only considers spatial features, without considering temporal relations between images. We hypothesize that {\bf generalization of transfer learning can be enhanced by leveraging spatial and temporal features in cross-domain training}. 
Table~\ref{tab:table4} 
shows results where the {\em resulting trained parameters from the neural network incorporate around 78\% of the global spatio-temporal features}. 

\vspace*{0.5em}
\noindent
\textbf{Key Insights.} Our two main contributions towards current transfer learning frameworks are (1) the transfer of spatio-temporal information from both the network model and the image sequences, as opposed to solely spatial information; and (2) dynamics-aware feature transfer by complementing training data with saliency, gradient, and edge maps. We describe our enhanced transfer learning method in two phases: the first of which comprises  training in the source domain and the second comprises training with transferred features from Phase One, along with the extracted spatio-temporal features in saliency maps in Phase Two. 

\subsection{Our Transfer Learning Method}
        
We propose to adopt a CNN+LSTM model to improve the generalization of transfer learning across domains, by better exploiting the spatio-temporal features at both the {\em network level} and the {\em data level}, as shown in Fig.~\ref{fig:experiment-diagram}. 

\textbf{Learning Task. }
We train and test our method on a binary classification task, where an entire image sequence will be labeled as “collision” or “safe”. These labels refer to the outcome of the image sequence; in sequences which start out safe and end with a collision, the label for the entire sequence will be “collision”. While the task seems simple, identification of a potential collision in a video sequence will need temporal information to classify well. For example, vehicle velocity cannot be captured solely by an independent image. In the case of an intersection, some vehicles may have high velocities in certain frames which will change the outcome of the classification. 

\textbf{Datasets. }
We collected training data from the CARLA simulator {CARLA} (version 0.9.5), where a human user controls the vehicle and a background autopilot API script formats and records images and labels. 
Data is collected in image sequences, where one example represents 15 images in chronological order, stacked. 
Images are of size 420 by 280, making the resulting input dimension to the model (N, 15, 3, 420, 280), where N represents the minibatch. 
A total of about 6,000 examples were used for training, and were collected from the CARLA Town 3 environment. 

While there are many other well-established datasets for autonomous driving tasks~\cite{Camvid, kitti, torcs, cityscapes, synthia, gtaV, davis, waymo, audi_dataset, honda, apolloscape, cadcd, nuscenes, bdd100k}, we find that these datasets very rarely samples representing collision scenarios at all, and are not best choices for 'collision classification'. 

We test our transfer learning model on a variety of test datasets in order to show its ability to generalize to unseen domains. 
The first test dataset we use is data from Town 1 in CARLA. This represents the same visual domain, but a different environment.
In order to test in the real world, we collected dashcam footage of car collisions from YouTube videos and combined those collision sequences with standard driving data from Audi ~\cite{audi_dataset} to create a real-world dataset. This highlights the main advantage of the collision classification task; with real-world collision data being difficult to obtain, having binary labels makes it possible to leverage real-world data from the internet without having to go through collection of costly vehicle accident data. Similarly, we use data from the DeepDrive simulator ~\cite{DeepDrive} to observe performance in other virtual domains.

While there may be differences across datasets in terms of vehicle dynamics and image quality, the hope is that the transfer learning method will learn to generalize domain-specific details. One problem with driving datasets in general is that there is no specific standard to the collection of these datasets regarding how data is collected. Because of this, scraped YouTube data can be a great example of a case where models can use crowd sourced data to learn, where each contributing video sequence is from a different person with varying qualities of dash cameras.

\textbf{Model Architecture. } Our model architecture consists of two convolutional layers, two inception modules~\cite{Szegedy_Liu_Jia_Sermanet_Reed_Anguelov_Erhan_Vanhoucke_Rabinovich_2015}, two LSTM layers, three fully connected layers, and a softmax layer. This model is based on a collision-detection model proposed in~\cite{CNN-LSTM-towardsdatascience}. When training from scratch, we initialize the hidden states of the LSTM with random noise, as well as initialize the other weights with standard Xavier initialization. 
    
\begin{figure*}[t!]
    \centering
    \includegraphics[width=17.7cm,height=3.5cm]{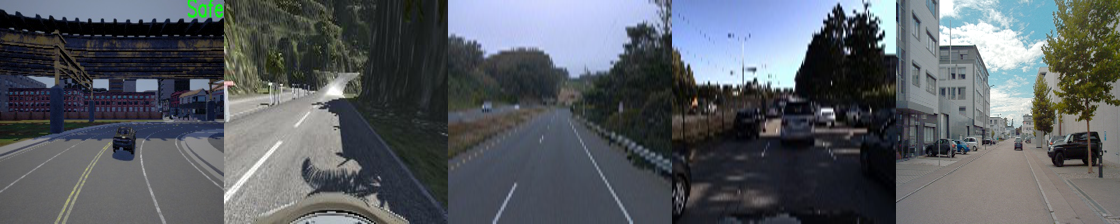}
    \vspace*{-2em}
   \caption{Examples from each test dataset (Left to Right): CARLA~\cite{CARLA}, Udacity~\cite{udacity}, Waymo~\cite{waymo}, Honda~\cite{honda}, Audi~\cite{audi_dataset}. While each dataset has varying levels of coloration, quality, resolution, etc., we can observe that the general scene layout from the dashcam point of view remains consistent. }
    \vspace*{-1em}
    \label{fig:dataset-examples}
\end{figure*}

\begin{figure*}[t!]
    \centering
    \includegraphics[width=17.7cm,height=3.5cm]{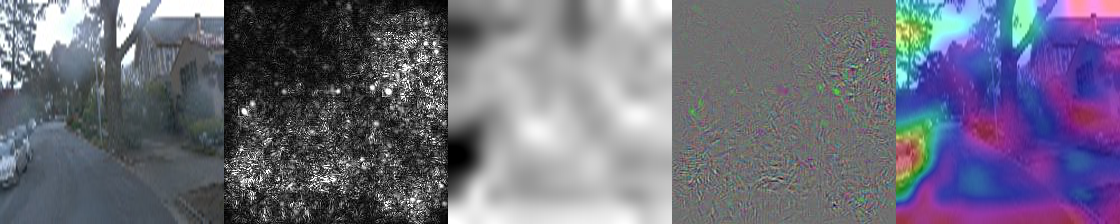}
    \vspace*{-2em}
   \caption{Example of gradient and saliency maps used in data augmentation (Left to Right): Original image (Waymo), Grayscale vanilla backpropagation saliency, Grayscale GradCAM~\cite{gradcam} map, RGB vanilla backpropagation saliency, RGB GradCAM map overlayed on top of the original image. Only the grayscale versions of gradient and saliency maps were used to augment the dataset; the colored versions are shown for visualization purposes. }
    \vspace*{-1.5em}
    \label{fig:saliency-gradient-examples}
\end{figure*}

\textbf{Phase 1 Training. } 
Phase 1 training represents a standard from-scratch training of the network described above. The learning task is the binary classification task of recognizing dangerous from safe scenarios in video. We train this network on data collected from the CARLA simulator~\cite{CARLA}, e.g. Town 3, as shown in the upper box of Fig.~\ref{fig:experiment-diagram}. 


\textbf{Salient Data Generation.}
After Phase 1, we generate GradCAM~\cite{gradcam} maps, Vanilla Backpropagation Gradient maps, and edge maps produced by Canny Edge Detection~\cite{Canny_1986} based on target domain data and use it to complement the transfer learning process. GradCAM and Vanilla Backpropagation maps are generated using an AlexNet model that is pre-trained on ImageNet~\cite{imagenet}. 
We use gradient/saliency maps from a larger, pre-trained ImageNet model to hint the Phase 2 network with bigger-picture context of the source and target domain data. While another option is to generate gradient and saliency maps using the Phase 1 network, this would be repetitive with respect to the transferred weights. To achieve this, we use a convolutional network visualization code library provided by ~\cite{uozbulak_pytorch_vis_2021}.
We generate saliency and gradient maps for a subset of the source domain data. This subset ratio can be adjusted. For our experiments, we randomly generate maps for 10\% of our target domain data.
The resulting saliency and gradient maps are used to augment the target domain dataset during training of Phase 2. In other words, we use the resulting 3-channel image representations of the maps to combine with the target domain training data as input to the Phase 2 model. 


\textbf{Phase 2 Training. }
In Phase 2, the trained weights from Phase 1 are transferred using weight initialization to Phase 2, which involves the same architecture as Phase 1. Both CNN weights and LSTM hidden states (i.e. model parameters and weights) are transferred. Saliency and gradient maps are generated from a pre-trained ImageNet model, then concatenated onto the target domain images (i.e., added as auxiliary image channels) for training in Phase 2. As mentioned before, the motivation for using these maps are to help the Phase 2 model gain context of driving scenarios from a network that recognizes a large range of objects. Reframing in another sense, {\em we transfer knowledge from a network with more diverse classes by adding features maps to the training data}, rather than through weight transfer. 
%
%
Training then proceeds as usual using data from our target domain data, Town 1 of CARLA. 



\subsection{Measuring Similarity Between Datasets}
The goal of our transfer learning method is to learn relevant driving features from the source domain data in Phase 1 and to transfer those features for use in Phase 2, which is then trained on target domain data. It would be useful to have an indicator or a similarity metric to estimate the effectiveness of transfer learning. In our experiments, we include measurements on mean cosine similarity between the feature representation of images. Specifically, we extract the outputs of the inception modules in the CNN architecture for inputs from each dataset, and calculate the cosine similarity between these output vectors. The calculation of cosine similarity is defined below as: 
\begin{align*}
    cos\_sim(A,B) &= \frac{A \cdot B}{\|A\| \times \|B\|}
\end{align*}
where $A$ and $B$ are output vectors of the inception module with image inputs from the source domain dataset and the target domain dataset, respectively. 


Cosine similarity is further used in an ablation study on temporal dynamics in driving data to understand the sensitivity of model outputs to movements across frames. For more results on this study, see Table~\ref{tab:table4}, and Tables~\ref{tab:table5},~\ref{tab:table6}, and ~\ref{tab:table7} in the Appendix.

\begin{table*}[h!]
    \centering
    \begin{tabular}{c|c c c c c}
    \toprule
        & \multicolumn{5}{c}{\bf Test Accuracy on Unseen Datasets (\%) $\uparrow$} \\
        Model & CARLA Town2 & DeepDrive &  Audi+YT & Honda+YT & Youtube  \\
        \midrule
        Mean Cosine Similarity (to CARLA Town1) & 0.77 & 0.697 & 0.623 & 0.591 & 0.55 \\
        \midrule
        Baseline Model~\cite{akhaurizhenglin2020} & 71 & 65 & 63 & 60 & 59 \\
          Our Method without Data Aug & 88 & 80 & 75 & 73 & 76 \\
        Our Method & \textbf{91} & \textbf{85} & \textbf{82} & \textbf{82} &\textbf{81} \\
        \midrule
        CNN+LSTM w/ txf weights $\Delta$ Improvement (\%) & 23.94   & 23.07   & 19.04 & 21.67  &  28.81 \\
        Ours Overall $\Delta$ Improvement (\%) & {\bf 29.58} & {\bf 30.77} & {\bf 30.15} & \textbf{36.6} & {\bf 37.29} \\
        \bottomrule
    \end{tabular}
    \caption{{\bf Comparison of performance accuracy (\%) between the baseline model and our model across unseen test datasets (Higher better $\uparrow$).} Unseen test dataset include: 1) CARLA Town 2, which represents the same visual domain but a different environment, 2) DeepDrive~\cite{DeepDrive}, another popular driving simulator to represent a different virtual domain, and 3) real-world data collected from Audi~\cite{audi_dataset} and YouTube collision videos. 
    Both our model and the baseline are trained on Carla Town 1 dataset. We measure the semantic differences in each domain using cosine similarity; higher values indicate closer semantic relationships to the training dataset, and serve as a proxy indicator of transferability of learning across domains. We also compare our model with only transferred weights to our model overall, which involves transferred weights and saliency/gradient/edge maps. Between these two results, we observe that saliency, gradient, and edge maps can significantly improve test performance. 
    {\bf Overall, our model shows better generalization across environments} on which it was not trained by {\bf 29.58\% up to 37.3\%} improvement in accuracy over the baseline model~\cite{akhaurizhenglin2020}. }
    \vspace*{-2em}
    \label{tab:results}
\end{table*}

\begin{table*}[h!]
    \centering
    \begin{tabular}{c|c c c c c c c c}
    \toprule
        & \multicolumn{5}{c}{\bf Mean Error in Steering Angle Prediction (\%) $\downarrow$} \\
        Model & CARLA Town2 & DeepDrive  & Udacity & Audi & Waymo & Honda \\
        \midrule
        Baseline Model~\cite{akhaurizhenglin2020} &3.21 & 3.8 & 5.2 & 6.3 & 7.2 & 6.9 \\
          Our CNN+LSTM without Salient Data & 2.8 & 3.1 & 4.6  & 5.8 & 6.6 & 6.2\\
        Our CNN+LSTM Overall & \textbf{1.9} & \textbf{2.9} &\textbf{4.3} & \textbf{5.6} & \textbf{6.1} & \textbf{5.9} \\
        \midrule
        
        Ours Overall, $\Delta$ Improvement (\%) & {\bf 40.8} & {\bf 23.7} & {\bf 17.3} & {\bf 11.1}  & \textbf{15.3} & \textbf{14.5} \\
        \bottomrule
    \end{tabular}
    \caption{\textbf{Mean error for end-to-end steering prediction task for Sim2Sim and Sim2Real (Lower better $\downarrow$).} This table shows a performance comparison between our model and other baselines trained on simulator data (Udacity) for both Phase 1 and Phase 2. We tested these models against several simulated and real-world test datasets, most of which are from visual domains not seen during training. We calculate $\delta$ improvement (\%) by taking the difference between mean error of our method and the baseline model, then normalizing the difference with the result of the baseline model.  Ours achieves up to {\bf 40.8\%} improvement in sim2sim and up to {\bf 23.7\%} on previously unseen datasets for the steering prediction task over the baseline model. }
    \vspace*{-2em}
    \label{tab:results-steering}
\end{table*}

        
        

\begin{table*}[h!]
     \centering
     \begin{tabular}{c|c c  }
     \toprule
         Seq & Scenario for spatio-temporal features measurement & Mean Cosine  \\
         \midrule
         1 & Agent moving with nearby vehicles  & \textbf{0.7983} \\
         2 & Agent moving on empty roads  & \textbf{0.8887} \\
         3 & Agent turning & \textbf{0.732} \\
         4 & Agent stationary with dynamic obstacles around & \textbf{0.712} \\
         \midrule
         $\mu$ & All &  \textbf{0.7827}\\
     \bottomrule
     \end{tabular}
     \caption{{\bf A scenario-based case study on the role of spatio-temporal features:} It includes agent's orientation, optical flow of the edge information across frames, dynamic obstacles and moving vehicles using the approach described for the ablation study. We also perform a study on the combination of the role of the four global spatio-temporal features including optical flows, rotational components, gradient flows and dynamic obstacles. We observe that these features together account for about 78\%, while other spatial features in the background contributing 22\%, towards the agent's decision making. }

     \label{tab:table4}
     \vspace*{-3em}
 \end{table*}

\section{Experiments \& Results}
\label{sec:methodology}


\subsection{Evaluation on Collision Classification}
We evaluate our transfer learning method with the collision classification task by comparing between different models and datasets. We choose our baseline transfer learning model to be that of Akhauri et al.~\cite{akhaurizhenglin2020}, which aligns with the goal of handling collision scenarios with transfer learning. However, the architecture of the baseline model does not take into account temporal features. Specifically, our model has the addition of LSTM layers as well as knowledge (i.e. salient features) transfer on the data level and hidden states initialization at the network level. 
 
We compare our model to a similar version of the transfer learning process which does not involve transfer of LSTM hidden states, but only of the CNN weights. In this way, we measure the effect of transferring information between LSTM hidden states across models. 

In addition, we also compare our model's performance across different unseen test datasets. We show that the performance improvement of our model can be generalized to datasets of varying quality and style, as well as in both virtual simulator domains and real world. Even if a model is tested on an unseen dataset with high visual contrast to the training dataset, our model shows improvement in generalizing features learned from training compared to other baselines. 

Our transfer learning model, which is trained solely on CARLA Town 1 and Town 3, shows classification accuracy improvement on CARLA Town 2. We also show test accuracy on datasets from DeepDrive~\cite{DeepDrive}, Audi~\cite{audi_dataset}, Honda~\cite{honda}, and YouTube. As mentioned from the transfer learning section before, we combine these datasets with video frames depicting collision scenarios from crowdsourced YouTube videos. 
Results for this evaluation can be found in Table~\ref{tab:results}.

\subsection{Key Observations and Discussion}

We observe the following key findings in Table~\ref{tab:results}:
\begin{itemize}
\item The mean cosine similarity metric correlates linearly to the performance improvement in using our dynamics-aware transfer learning framework.  Higher metric values correspond to higher similarity in the visual domains, thus benefiting less from the enhancement as expected given the base model should work well in similar domains. 
However, {\em where the visual domains have significantly more disparity, such as in the case of `sim-to-real', our enhanced transfer learning framework provides better generalization across domains (up to 37\%)}.
\item In visual domains with high similarity scores, the use of LSTM, along with the transfer of its hidden states (i.e. the model parameters or weights), can adequately capture the common spatio-temporal features in the model itself, thus sufficiently improve the transferability of the network training, as seen in `sim-to-sim' case for CARLA Town 1 and 2 ({\em or} Town 1 and DeepDrive).
\item In visual domains with low similarity scores, the LSTM hidden states may not capture spatio-temporal features as well, the introduction of saliency, edges, and gradient maps together provide additional spatio-temporal information not already embedded in the network model, as in the `sim-to-real' case between CARLA Town 1 and a collection of Youtube videos ({\em or} Audi driving data).
\end{itemize}

\subsection{Evaluation on End-to-End Steering}
Our transfer learning method can also be generalized to other autonomous driving tasks, such as end-to-end steering. End-to-end steering can be described as a regression problem in which image data are inputs to the model, and the output is a scalar value prediction of the steering angle representing safe driving. Steering can benefit from transfer of spatio-temporal features, as there is an inherent relationship from frame to frame describing the velocity of the vehicle. 

For evaluation on end-to-end steering, we keep the same architecture as that from the collision classification task. Thus, we achieve results on this task via behavioral cloning, and measure the deviation from the ground truth steering angle to the predicted steering angle, in degrees.
We train Phase 1 on the CARLA dataset, and Phase 2 on the Udacity dataset. 
We evaluate our model on several well-known driving datasets with steering angle labels: Udacity~\cite{udacity}, Audi~\cite{audi_dataset}, Waymo~\cite{waymo}, and Honda~\cite{honda}. We also evaluate performance on the datasets we used previously in the collision classification task, CARLA~\cite{CARLA} and DeepDrive~\cite{DeepDrive}. 
We observe considerable error reduction up to {\bf 40.8\% in sim2sim} and up to {\bf 23.7\% on previously unseen real-world driving datasets} for the steering prediction task over the baseline model.

\begin{figure}[t!]
    \centering
    \includegraphics[width=7cm,height=5cm]{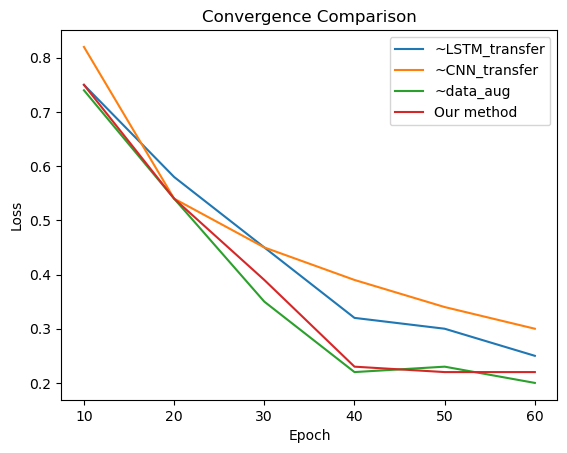}
    \vspace*{-1em}
       \caption{\textbf{Comparison of convergence on Phase 2 model between different transfer learning methods.} We use the tilde $\sim$ notation to denote that this particular part of the transfer learning process is missing. For example, $\sim$LSTM\_transfer denotes the model which does not involve transfer of hidden states from Phase 1 to Phase 2. From this line plot, we observe that neither CNN or LSTM transfer alone converges faster than our method, which combines both. In addition, data augmentation does not appear to affect the convergence process. This is not a surprising result, since data augmentation occurs from the dataset, and does not cause weights to optimize any faster. }
    \vspace*{-1em}
    \label{fig:convergence-examples}
\end{figure}
\section{Conclusion and Future Work}
In this work, we present two main contributions: 1) transfer of spatial-temporal features through transfer of LSTM hidden state embeddings and CNN weights using a CNN+LSTM model architecture for vision-based autonomous driving networks, and 2) dynamics-aware spatio-temporal feature extraction to complement regular image data with saliency, gradient, and edge maps for training, where the model perception of visual elements using cosine similarity is used to select these features.

Beyond LSTMs, transfer learning for autonomous driving may benefit from the use of Transformer models. Even though transformer models suffer from limited data sequence length, they are much easier to train due to the availability of pre-trained models and are more interpretable than LSTM models due to their attention mechanism. For transformers, it may be more intuitive to identify important spatio-temporal features not only in short-term memory, but also long-term memory. 

Formalizing vehicle dynamics is also difficult on the network level. While it is convenient to have end-to-end systems as an all-in-one solution, it loses the benefit of modular systems where inputs and outputs are easily controlled. With general machine learning models suffering from bias problems, it would be helpful to define priors or hints for end-to-end models during the training process. 
More directly, contributions of this work can be extended to more complex learning tasks, such as multi-class classification, segmentation, or steering angle prediction. 
The addition of salient data (i.e. saliency maps, gradient maps, edge maps) process can be streamlined as well.  While it requires one more phase of training to complete, it also greatly helps generalization in unseen domains. Generation of such data can be streamlined in online learning, rather than our method of offline learning. 





\bibliographystyle{IEEEtran}
\bibliography{references}


\clearpage
\newpage 

 \begin{table}[h!]
     \centering
     \begin{tabular}{c|c c  }
     \toprule
         Seq & Agent orientation(wrt the lane center) & Mean Cosine  \\
         \midrule
         1 & 70-80  & \textbf{0.3454} \\
         2 & 50-60  & \textbf{0.32} \\
         3 & 30-40 & \textbf{0.2908} \\
         4 & 5-20 & \textbf{0.159} \\
         \midrule
         $\mu$ & All &  \textbf{0.277}\\
     \bottomrule
     \end{tabular}
     \caption{{\bf Quantitative analysis on impact of the agent's orientation across frames in the latent representations using cosine similarity metric.} We observe that the agent's change in orientation as a spatio-temporal feature accounts for 28\% contribution in the hidden states of the trained model.}

     \label{tab:table5}
     \vspace*{-3em}
 \end{table}

  \begin{table}[h!]
     \centering
     \begin{tabular}{c|c c  }
     \toprule
         Seq & Scenario for Optical Flow temporal feature measurement & Mean Cosine  \\
         \midrule
         1 & Agent moving with nearby vehicles  & \textbf{0.48} \\
         2 & Agent moving on empty roads  & \textbf{0.40} \\
         3 & Agent turning & \textbf{0.54} \\
         4 & Agent stationary with dynamic obstacles around & \textbf{0.42} \\
         \midrule
         $\mu$ & All &  \textbf{0.46}\\
     \bottomrule
     \end{tabular}
     \caption{{\bf Quantitative analysis on impact of the optical flows across frames in the network latent representations using cosine similarity metric.} We observe that optical flows as a spatio-temporal feature accounts for around 46\% contribution in the hidden states of the trained network.}

     \label{tab:table6}
     \vspace*{-3em}
 \end{table}

 \begin{table}[h!]
     \centering
     \begin{tabular}{c|c c  }
     \toprule
         Seq & Scenario for Dynamic obstacles feature measurement  & Mean Cosine  \\
         \midrule
         1 & Agent moving with nearby vehicles  & \textbf{0.18} \\
         2 & Agent moving on empty roads  & \textbf{0.13} \\
         3 & Agent turning & \textbf{0.14} \\
         4 & Agent stationary with dynamic obstacles around & \textbf{0.18} \\
         \midrule
         $\mu$ & All &  \textbf{0.157}\\
     \bottomrule
     \end{tabular}
     \caption{{\bf Quantitative analysis on impact of dynamic obstacles across frames in the network latent representations using cosine similarity metric.} We observe that dynamic obstacles as a spatio-temporal feature accounts for around 16\% contribution in the hidden states of the trained model.}

     \label{tab:table7}
     \vspace*{-3em}
 \end{table}

\end{document}